\documentclass[letterpaper]{article} 
\usepackage{aaai24}
\usepackage{booktabs}

\usepackage{float}
\usepackage{adjustbox}

\usepackage{times}  
\usepackage{helvet}  
\usepackage{courier}  
\usepackage[hyphens]{url}  
\usepackage{graphicx} 
\usepackage[numbers]{natbib}  

\usepackage{caption} 
\usepackage{algorithm}
\usepackage{algorithmic}
\usepackage{xcolor}
\usepackage{color}
\usepackage{url}

\usepackage{multirow}
\usepackage{makecell}
\usepackage{amssymb}
\usepackage{pifont}

\usepackage{newfloat}
\usepackage{listings}
\DeclareCaptionStyle{ruled}{labelfont=normalfont,labelsep=colon,strut=off} 
\floatstyle{ruled}
\newfloat{listing}{tb}{lst}{}
\floatname{listing}{Listing}
\usepackage{hyperref} 
\usepackage{verbatim}
\usepackage{markdown}

\title{SuperCLUE-Math6: Graded Multi-Step Math Reasoning Benchmark for LLMs in Chinese}
\author{
    Liang Xu, \ 
    Hang Xue,\ 
    Lei Zhu, \ 
    Kangkang Zhao \ 
}
\affiliations{
    SuperCLUE team

        contact@superclue.ai 

}

\usepackage{bibentry}

\begin{document}
\pagestyle{plain}

\maketitle

\begin{abstract}
We introduce SuperCLUE-Math6(SC-Math6), a new benchmark dataset to evaluate the mathematical reasoning abilities of Chinese language models. SC-Math6 is designed as an upgraded Chinese version of the GSM8K dataset with enhanced difficulty, diversity, and application scope. It consists of over 2000 mathematical word problems requiring multi-step reasoning and providing natural language solutions. We propose an innovative scheme to quantify the reasoning capability of large models based on performance over problems with different reasoning steps. Experiments on 
13
representative Chinese models demonstrate a clear stratification of reasoning levels, with top models like GPT-4 showing superior performance. SC-Math6 fills the gap in Chinese mathematical reasoning benchmarks and provides a comprehensive testbed to advance the intelligence of Chinese language models.~\footnote{Our benchmark can be found at \url{https://www.CLUEbenchmarks.com/superclue_math6.html}}.
\end{abstract}

\section{Introduction}

Recent advances in large language models like GPT-4 ~\citep{[1]}  have sparked great interest in evaluating their proficiency in solving reasoning problems. While benchmarks like GSM8K~\citep{[2]} have been influential, they are limited to English and do not sufficiently test multi-step inference. To overcome these limitations and systematically assess the mathematical reasoning of Chinese models, we introduce SC-Math6 as an upgraded Chinese version of GSM8K.

\begin{figure*}[h] 
\centering
\includegraphics[width=0.8\textwidth, height=0.48\textheight]{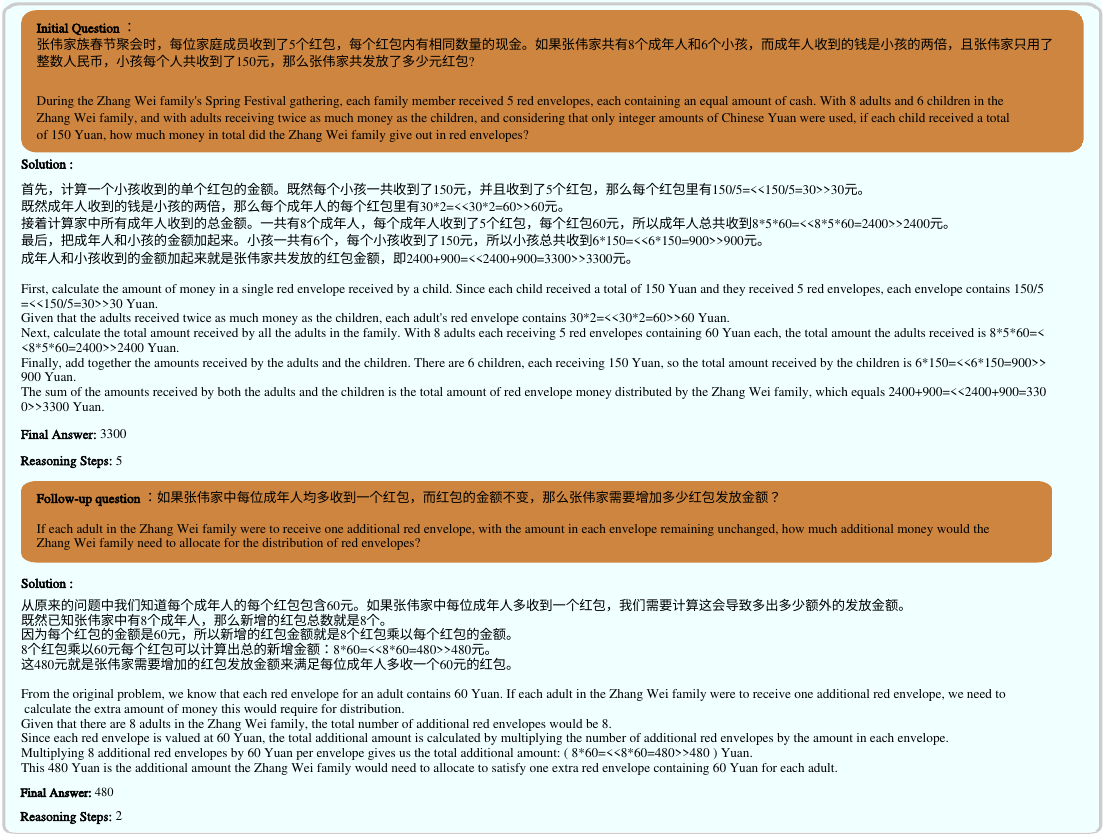}
\caption{An example of a problem in SC-Math6}
\label{fig:example_math}
\end{figure*}

SC-Math6 has 1072 unique problems covering a diverse range of grade school math topics. Each problem is presented in a native Chinese context and accompanied by a detailed natural language solution walkthrough. Moreover, SC-Math6 provides a follow-up question for each initial query to assess the model's continuous reasoning ability during interaction with users (As shown in Figure ~\ref{fig:example_math}. For more examples, please refer to the appendix \ref{appendix:appendix_a}). We also propose a novel scoring scheme that combines performance over problems with different reasoning steps and overall accuracy to produce interpretable and fair reasoning levels from 1 to 5. Disparities and Correlations of SC-Math6 and GSM8K is presented in Table ~\ref{tab:GSM_vs_Math6}

\begin{table}[H]

\centering
\begin{adjustbox}{scale=0.75}
\begin{tabular}{llcccc}
\toprule
\centering
\textbf{Comparison Item} & \textbf{SC-Math6} & \textbf{GSM8K} \\
\midrule
Mathematical Logic Reasoning & YES & YES \\
\hline
Natural Language Solutions & YES & YES \\
\hline
Elementary Mathematical Knowledge & YES & YES \\
\hline
Multi-step Reasoning & YES & YES \\
\hline
\textbf{Native Chinese Context} & YES & NO \\
\hline
\textbf{Multi-round In-depth Reasoning} & YES & NO \\
\hline
\textbf{Reasoning Steps in Problems} & YES & NO \\
\hline
\textbf{Interpretable Reasoning Level for LLMs} & YES & NO \\
\hline
Number of Test Questions & 2144 (1072 Pairs) & 1300 \\

\bottomrule
\end{tabular}
\end{adjustbox}
\caption{SC-Math6 and GSM8K: Disparities and Correlations}
\label{tab:GSM_vs_Math6}
\end{table}

Our experiments on 13 major Chinese models demonstrate a clear stratification of reasoning capabilities. Advanced models like GPT-4 exhibit remarkably high accuracy on multi-step problems, while lower-level models show large performance gaps. The diversified grading scheme provides a reference for model selection and evaluation. SC-Math6 thus contributes the first comprehensive Chinese benchmark to assess and improve the mathematical reasoning abilities of Chinese language models. 

\begin{figure}[H]
\centering
\includegraphics[width=0.5\textwidth, height=0.3\textheight]{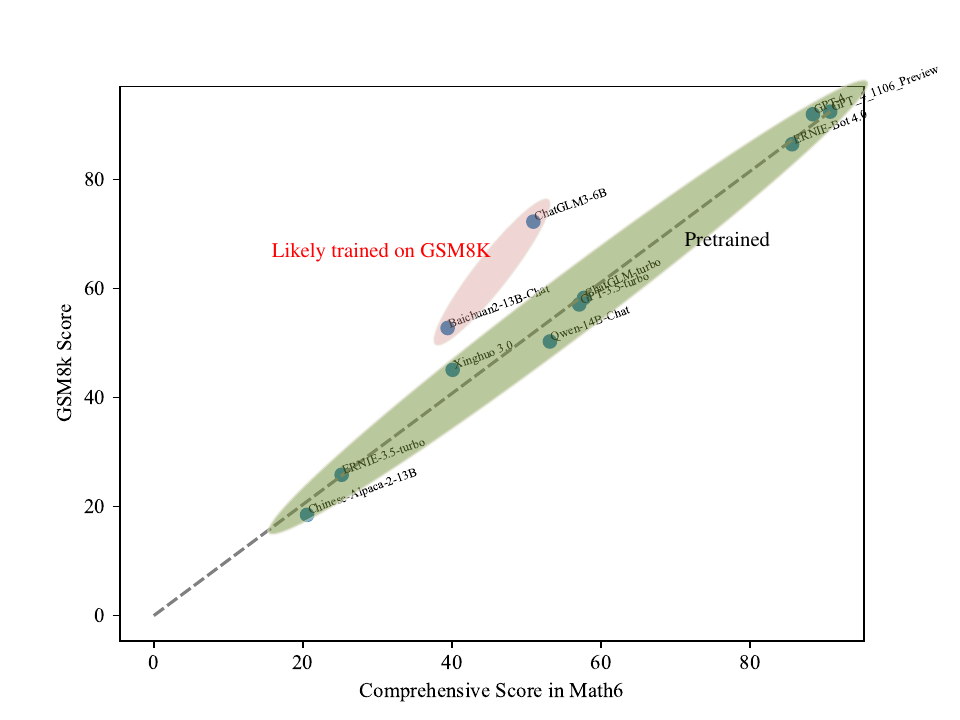}
\caption{Trend analysis between GSM8k and SC-Math6. SC-Math6 aligns with GSM8K yet demands in-depth reasoning, whereas some models may struggle with SC-Math6.}
\label{fig:GSM8k_Math6}
\end{figure}

This work pioneers the systematic evaluation and benchmarking of mathematical reasoning capabilities of major Chinese language models. The key contributions are three-fold:

First, the construction of SuperCLUE-Math6, the first native Chinese multi-turn, multi-step mathematical reasoning dataset for assessing model logical thinking and reasoning skills.

Second, the proposal of a novel transparent and consistent framework to parse and evaluate model reasoning levels, providing quantifiable metrics of model intellectual capabilities.

Third, comprehensive benchmarking and analysis of leading Chinese models on SuperCLUE-Math6, offering valuable insights into current model strengths, weaknesses and factors impacting reasoning performance.

Overall, this research fills the gap in Chinese mathematical reasoning evaluation and establishes an important benchmark for advancing the reasoning abilities of Chinese language models. The benchmark and insights lay a solid foundation for developing models with more human-like reasoning.

\section{SuperCLUE-Math6}

\subsection{Data Collection}

We first curated a large pool of Chinese math problems from elementary school exams and books and altered it manually to ensure it was unique. Two criteria were then applied to select problems requiring: {1) At least one step of reasoning, and 2) Error-free natural language solutions. This yielded 1072 unique arithmetic problems.

To evaluate the model's proficiency in sustained inferential reasoning throughout the interactive engagement, we designed multi-turn follow-up questions for each problem, bringing the total size to 2144.

Quality Control and Inspection of Questions. In the second round process, all questions were subjected to manual verification, which required the annotators to solve the questions themselves and record their answers. These were then compared with the provided reference answers and solution steps. If any inconsistencies were discovered, the problem was pinpointed and corrected. Corrections were made if there were issues with the answer or solution steps; if the question itself was ambiguous, it required clarification. Once consistency was confirmed, the process advanced to the next question.
After the manual verification, a final round of random sampling checks was conducted. Out of 50 pairs of questions, 1 pair was found to potentially have ambiguities and needed to be corrected, giving us a sampling accuracy of 98\%.

 The distribution of reasoning steps is controlled to prevent biases and test varied capabilities: 15-20\% with 1 step, 15-20\% with 2 steps, 45-50\% with 3 steps, and 5-10\% with 4-5 steps. The textual lengths of problems and solutions also exhibit high variability.

\subsection{Scoring Scheme and Reasoning Level}

To produce interpretable and fair quantification of reasoning capabilities, we propose a scoring scheme that combines:

\begin{itemize}
\item Reasoning Steps Score: Higher weight assigned for more steps based on the insight that longer reasoning chains are more difficult. Firstly, we separately compute the average score of the model corresponding to each inference step across the set of problems. Subsequently, we employ the number of inference steps as weights to calculate a weighted average of the scores at each step, thereby deriving a score that is weighted according to the number of inference steps. 
\item Overall Accuracy Score: The Overall Average Score is derived as the mean value of the Mean Accuracy and the Strict Interaction Accuracy. Mean Accuracy is computed by considering each question and its corresponding follow-up question as two separate items, thus calculating the average accuracy across 2144 questions. Conversely, Strict Interaction Accuracy is calculated by treating the question and its follow-up question as a unified interactive pair, with a point awarded only if both the question and the follow-up question are correctly answered, demonstrating proper reasoning. This Strict Interaction Accuracy is evaluated over a cohort of 1072 test pairs to establish the average interaction accuracy. 
\item Comprehensive Score: Calculated as the weighted sum of the Reasoning Steps Score and the Overall Accuracy Score, each component contributing equally to the final score.
\item Reasoning Level: The Reasoning Level of a language model is based on the Comprehensive Score, with levels ranging from 1 to 5, where level 5 is the highest and level 1 is the lowest. A threshold of 5 points is used to determine the levels. If the composite scores of the two models differ by less than 5 points, they are considered to be within the same level. This provides a transparent system to classify model capabilities.
\end{itemize}

The accuracy score is a commonly used evaluation metric to quantify the mathematical reasoning abilities of language models, but it fails to account for the varying difficulty levels of individual questions. Solving a more challenging problem should be awarded more points than solving an easier one, and the number of reasoning steps involved usually correlates with the difficulty level. Therefore, we incorporate the Reasoning Steps Score into our assessment framework. 

The advantage of the Reasoning Step Score lies in its ability to account for the varying difficulty levels of different questions. This score is validated by manual problem-solving, ensuring a high level of accuracy in the reasoning steps involved. Given that mathematical problems may have multiple solution methods, each with a potentially different number of reasoning steps, the reasoning step count is not necessarily unique. Therefore, the calculation of Reasoning Step Scores cannot eliminate the discrepancies in weighted precision.

On the other hand, the Overall Accuracy Score, while not considering the difficulty level of each question, ensures fairness and avoids bias that might be introduced during the weighting process. Therefore, we have not completely discarded the Overall Accuracy Score. Instead, we employ a weighted summation of Reasoning Steps Score and Overall Accuracy Score to calculate the unified score. This method balances the precision given by the Reasoning Steps Score with the fairness ensured by the Overall Accuracy Score, aiming to provide a more comprehensive evaluation of mathematical reasoning performance.

\subsection{Experiments and Analysis}
We evaluated 13 major Chinese models on SC-Math6 covering capacities from 13B to Proprietary APIs. Table ~\ref{tab:model_overall} presents the overall accuracy, Reasoning Steps Score, Comprehensive Score, and resulting Reasoning Level. Information on models is shown in Table~\ref{tab:model_information}.

\begin{table}[H]

\centering
\begin{adjustbox}{scale=0.60}
\begin{tabular}{lccccc}
\toprule
\centering
\textbf{Model Name} & \textbf{R Level} & \textbf{Comp. Score} & \textbf{Reas. Steps Score} & \textbf{OvrAcc Score} \\
\midrule
GPT-4-1106-Preview & \textbf{5} & 90.71 & 91.65 & 89.77 \\
GPT-4 & \textbf{5} & 88.40 & 89.10 & 87.71 \\
Ernie-bot 4.0 & \textbf{5} & 85.60 & 86.82 & 84.38 \\
GLM-4 & \textbf{5} & 84.24 & 85.72 & 82.77 \\
Xinghuo 3.5 & \textbf{5} & 83.73 & 85.37 & 82.09 \\
ChatGLM-Turbo & \textbf{4} & 57.70 & 60.32 & 55.09 \\
GPT-3.5-Turbo & \textbf{4} & 57.05 & 59.61 & 54.50 \\
Qwen-14B-Chat & \textbf{4} & 53.12 & 55.99 & 50.26 \\
ChatGLM3-6B & \textbf{3} & 40.90 & 44.20 & 37.60 \\
Xinghuo 3.0 & \textbf{3} & 40.08 & 45.27 & 34.89 \\
Baichuan2-13B-Chat & \textbf{3} & 39.40 & 42.63 & 36.18 \\
Ernie-3.5-turbo & \textbf{2} & 25.19 & 27.70 & 22.67 \\
Chinese-Alpaca2-13B & \textbf{2} & 20.55 & 22.52 & 18.58 \\

\bottomrule
\end{tabular}
\end{adjustbox}
\caption{SC-Math6 Model Reasoning Level. 'R Level' for Reasoning Level, 'Comp. Score' stands for Comprehensive Score, 'Reas. Steps Score' stands for Reasoning Steps Score, 'OvrAcc Score' stands for Overall Accuracy Score.}
\label{tab:model_overall}
\end{table}

\begin{table}[H]

\centering
\begin{adjustbox}{scale=0.80}
\begin{tabular}{llcccc}
\toprule
\centering
\textbf{Model Name} & \textbf{Organization} & \textbf{Access}  \\
\midrule
GPT-4-1106-Preview & OpenAI & API \\
GPT-4 & OpenAI & API \\
Ernie-bot 4.0 & Baidu & API \\
GLM-4 & ZhiPu & Web Page \\
Xinghuo 3.5 & Iflytek & API \\
ChatGLM-Turbo & ZhiPu & API \\
GPT-3.5-Turbo & OpenAI & API \\
Qwen-14B-Chat & Alibaba & API \\
ChatGLM3-6B & ZhiPu & Weight \\
Xinghuo 3.0 & Iflytek & API \\
Baichuan2-13B-Chat & Baichuan & Weight \\
Ernie-3.5-turbo & Baidu & Weight \\
Chinese-Alpaca2-13B & Yiming Cui & Weight \\

\bottomrule
\end{tabular}
\end{adjustbox}
\caption{Model information}
\label{tab:model_information}
\end{table}

Top models like GPT-4 exhibit remarkably high performance on multi-step problems, demonstrating advanced reasoning skills. There is also a clear stratification of capabilities, with higher-scoring models significantly outperforming lower ones. The diverse levels allow the selection of appropriate models based on application requirements.  \newline

We highlight several observations: \newline

1) Comparison between GSM8k Score and Comprehensive Score in SC-Math6

The comparative analysis of performance on the GSM8k and SC-Math6 benchmark. SC-Math6(Average model score is 55.34) aligns with GSM8K(Average model score is 59.21) yet demands more advanced reasoning, whereas models excelling on GSM8K may struggle on SC-Math6. It suggests that the SC-Math6 benchmark presents a greater level of difficulty. It was observed that, across the board, models tend to score lower on the SC-Math6 benchmark compared to GSM8k, with this trend being particularly pronounced for the ChatGLM3-6B and Baichuan2-13B-Chat models(As shown in Figure ~\ref{fig:GSM8k_Math6}).

2) Performance Declining during Multi-turn Interaction

In all models observed, the accuracy rate of the second iteration generally falls below that of the first, indicating a decline in model performance with increasing task complexity from the first to the second iteration(As shown in Table ~\ref{tab:Accuracy_multi_turn}). This trend is ubiquitous across all models, suggesting that special attention should be given to the stability and adaptability of models in sustained tasks during their design and optimization processes.

\begin{table}[H]

\centering
\begin{adjustbox}{scale=0.70}
\begin{tabular}{lccccc}
\toprule
\centering
\textbf{Model Name} & \textbf{Accuracy of Turn 1} & \textbf{Accuracy of Turn 2} & \textbf{Difference} \\
\midrule
GPT-4-1106-Preview & 95.43 & 89.37 & \textbf{-6.06} \\
GPT-4 & 94.12 & 87.13 & -6.99 \\
Ernie-bot 4.0 & 91.98 & 83.96 & -8.02 \\
GLM-4 & 90.39 & 83.02 & \textbf{-7.37} \\
Xinghuo 3.5 & 91.70 & 81.44 & -10.26 \\
ChatGLM-Turbo & 73.69 & 52.80 & -20.89 \\
GPT-3.5-Turbo & 70.99 & 53.59 & -17.40 \\
Qwen-14B-Chat & 73.23 & 46.26 & -26.97 \\
ChatGLM3-6B & 61.10 & 35.35 & -25.75 \\
Xinghuo 3.0 & 70.99 & 25.65 & \textbf{-45.34} \\
Baichuan2-13B-Chat & 58.86 & 33.77 & -25.09 \\
Chinese-Alpaca2-13B & 35.63 & 16.62 & -19.01 \\
Ernie-3.5-turbo & 43.00 & 20.43 & -22.57 \\

\bottomrule
\end{tabular}
\end{adjustbox}
\caption{SC-Math6 Accuracy during interaction}
\label{tab:Accuracy_multi_turn}
\end{table}

For instance, the GPT-4-1106-Preview model exhibited a first iteration accuracy rate of 95.43\%, which decreased to 89.37\% in the second iteration, marking an 6.06\% reduction in accuracy. Similarly, the ERNIE\_35\_Turbo model's accuracy rate declined from 43.00\% in the first iteration to 20.43\% in the second, constituting a 22.57\% decrease.\newline

3) Correlation between Instruction Compliance Ratio and Comprehensive Score

Models with high compliance to the instructed output formats also tend to achieve higher Comprehensive Scores, suggesting instruction understanding as an important indicator(As shown in Figure ~\ref{fig:instruction_compliance}). \newline

\begin{figure}[h]
\centering
\includegraphics[width=0.5\textwidth, height=0.3\textheight]{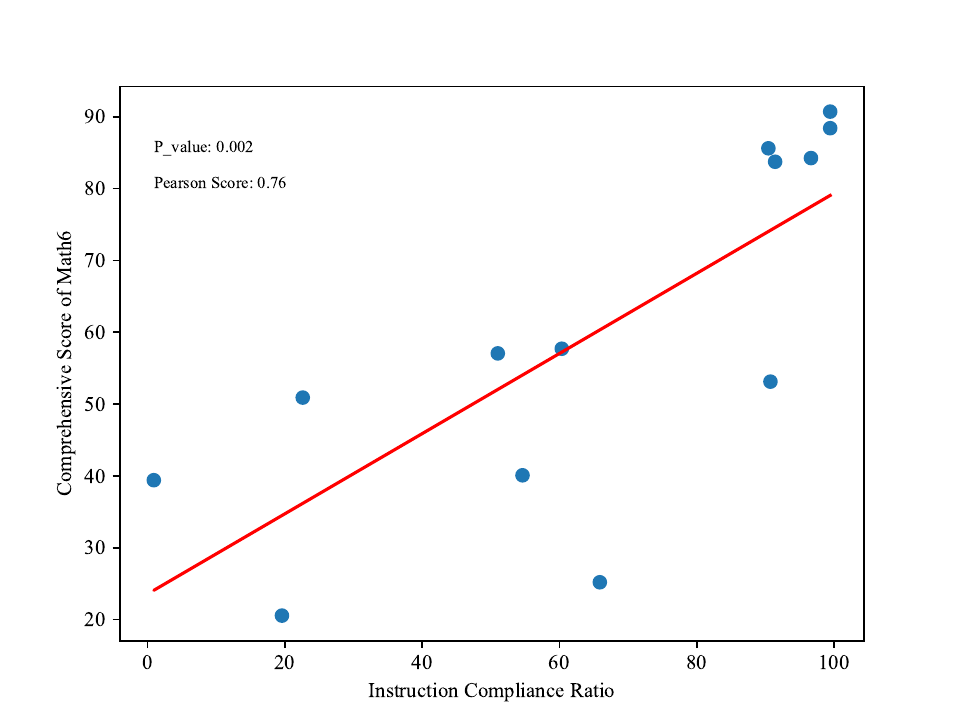}
\caption{Correlation between instruction compliance ratio and Comprehensive Score}
\label{fig:instruction_compliance}
\end{figure}

4) The Potential Relationship between Mathematical Reasoning Proficiency and Response Length

It has been observed that models yielding longer average response lengths tend to receive higher evaluation scores(As shown in Figure ~\ref{fig:better_performance_with_response_length_increasing}). In certain models, such as GPT-4-1106-Preview, a higher accuracy rate is accompanied by a longer average response length, which may suggest that these models are more precise when generating comprehensive responses. However, this trend is not consistently observed across all models, indicating that the relationship between response length and accuracy rate may be influenced by a multitude of factors, including the design of the model and the training data employed.

\begin{figure}[h]
\centering
\includegraphics[width=0.5\textwidth, height=0.3\textheight]{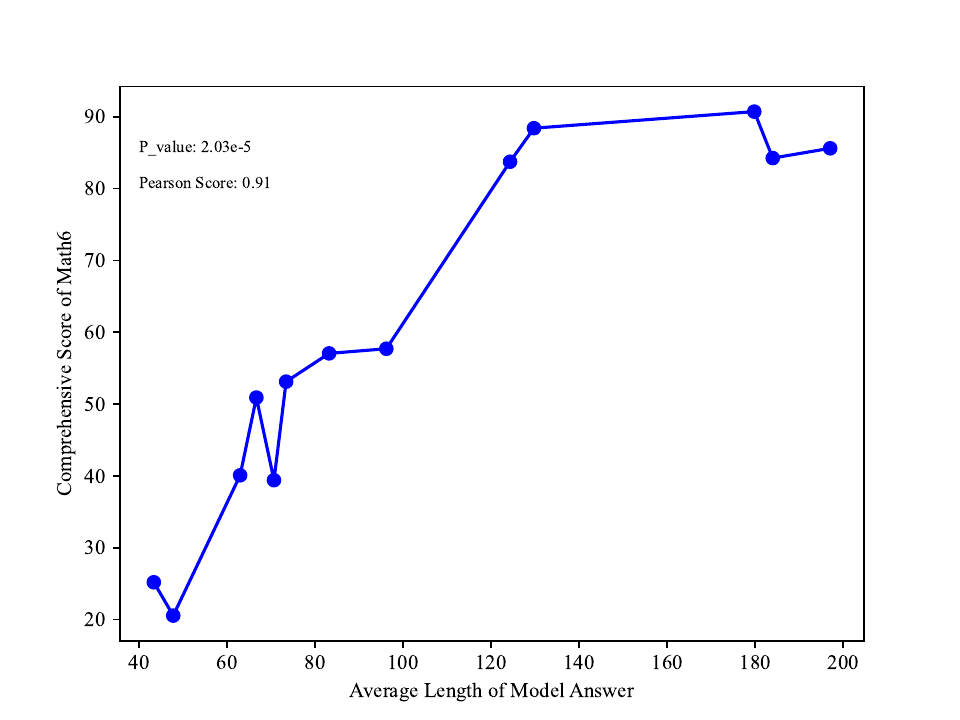}
\caption{Relation between response length and Comprehensive Score}
\label{fig:better_performance_with_response_length_increasing}
\end{figure}

The GPT-4-1106-Preview exhibits an average response length of 179.78, correlating with a higher accuracy rate, while the ChatGLM3-6B shows a comparatively shorter average response length of 66.66, with a corresponding lower accuracy rate. This implies that, in certain instances, there may be a correlation between response length and accuracy rate.  \newline

5) Performance Declining with Reasoning Steps Increasing
\begin{figure}[H]
\centering
\includegraphics[width=0.5\textwidth, height=0.3\textheight]{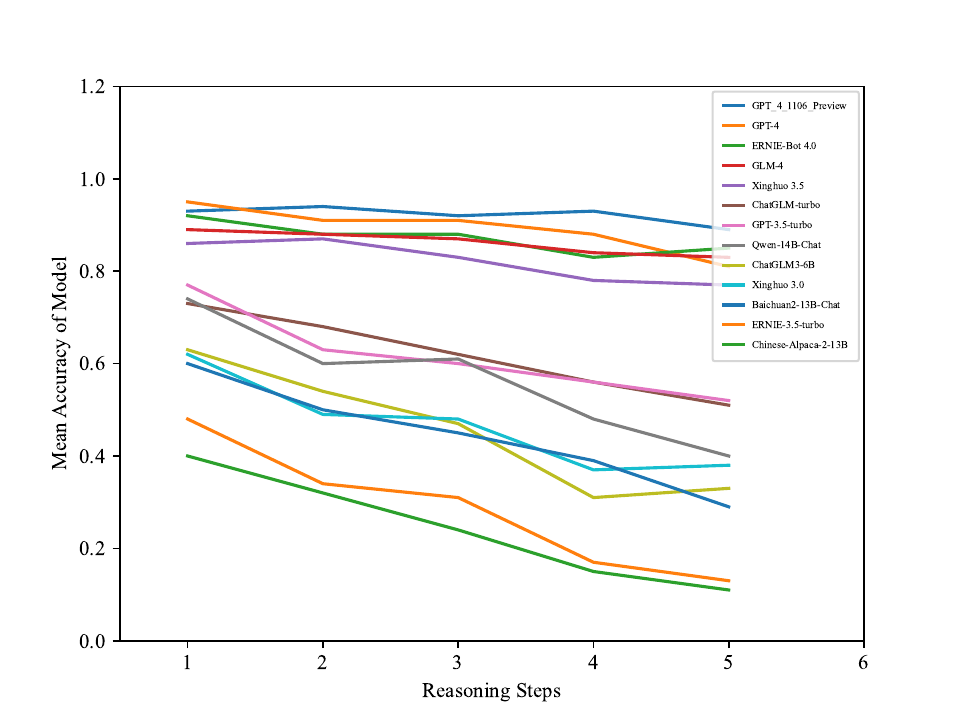}
\caption{Relation between reasoning steps and Mean Accuracy}
\label{fig:declining_performance_with_reasoning_steps_increasing}
\end{figure}

Analysis of the step-by-step scores reveals declining performance as problem complexity increases from 1 to 5 reasoning steps(As shown in Figure~\ref{fig:declining_performance_with_reasoning_steps_increasing}). This highlights the need to improve the model's capability to tackle more challenging problems requiring more reasoning steps. The results provide comprehensive insights to guide further progress on mathematical and general reasoning for Chinese language models. \newline


\section{Related Work}
Benchmarks to evaluate the reasoning skills of language models have gained increasing research attention. Existing datasets mostly focus on English, including GSM8K for mathematical reasoning~\citep{[2]}, MATH for complexity mathematical problem ~\citep{[3]}.  For general LLMs benchmark, we can find MT-bench ~\citep{[4]}, AlpacaEval ~\citep{[5]}. We can find reasoning benchmarks for NLP, such as  WinoGrad Schema Challenge for commonsense reasoning ~\citep{[6]}, and ARC for scientific question answering ~\citep{[7]}.
Our work aims to close the gap for the Chinese through a systematically designed mathematical reasoning benchmark. Our focus is to provide a benchmark to evaluate the general reasoning skills of Chinese language models. The diverse problems and reasoning patterns in SC-Math6 complement these methods to inspire new model designs and training strategies targeting enhanced mathematical intelligence.


\section{Conclusion}
We present SC-Math6 as the first native Chinese benchmark dataset for assessing the multi-step mathematical reasoning skills of language models in a multi-turn interaction. Developing human-like intelligence requires rich, diverse datasets like SC-Math6 that test sophisticated capabilities beyond pattern recognition. Our work aims to catalyze advances in Chinese models to better support real-world applications. The human-annotated natural language solutions also provide valuable data for training. We hope SC-Math6 will inspire exciting new model designs and training strategies targeting enhanced reasoning.












\appendix

\section{Examples for problem in SC-Math6}
Below are examples of problems in SuperCLUE-Math6.

\label{appendix:appendix_a}

\begin{figure*}[t] 
\centering
\includegraphics[width=0.75\textwidth, height=0.45\textheight]{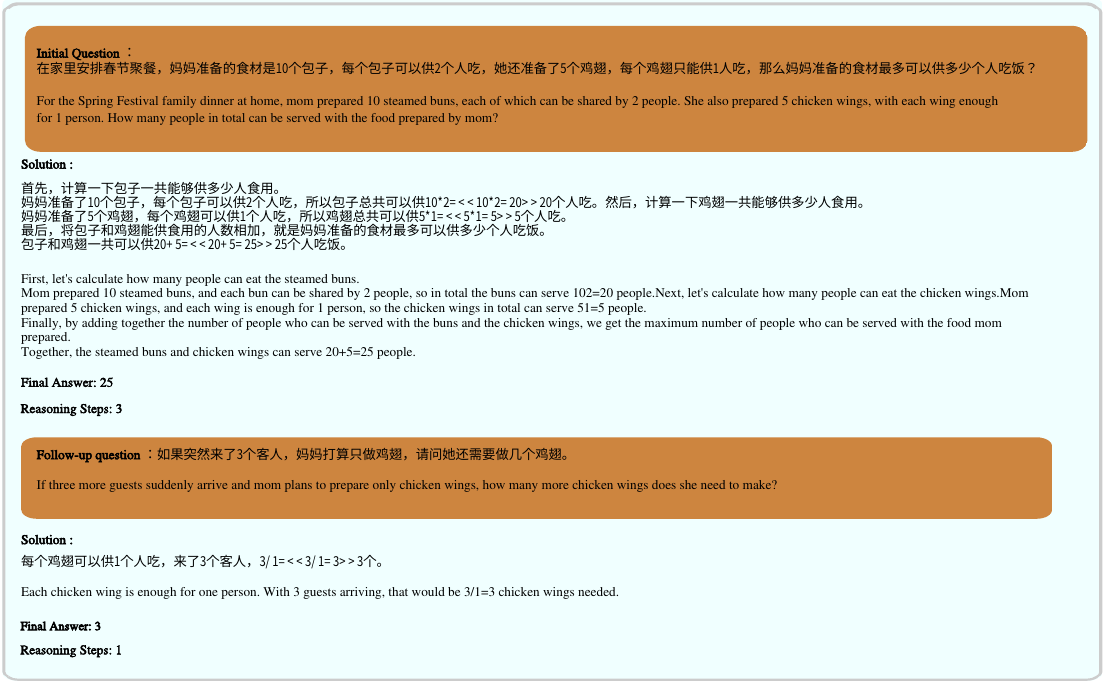}
\caption{An example of a problem in SC-Math6}
\label{fig:example_math_1}
\end{figure*}

\begin{figure*}[b] 
\centering
\includegraphics[width=0.75\textwidth, height=0.45\textheight]{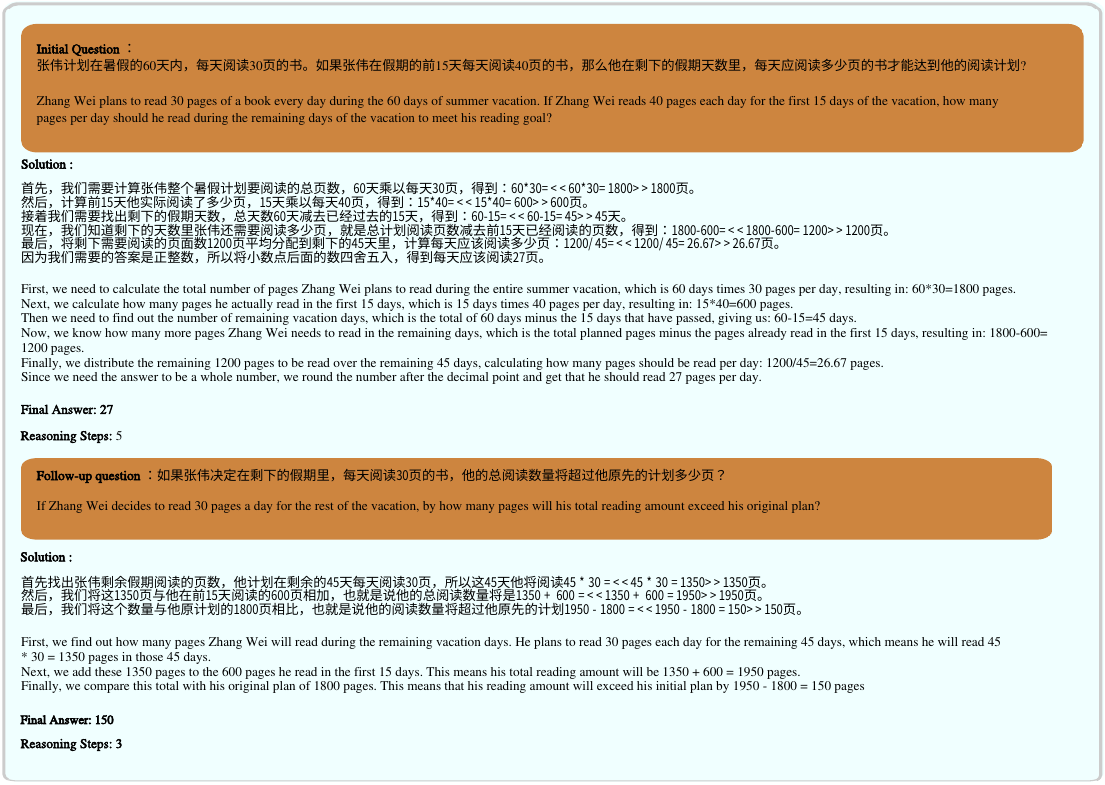}
\caption{An example of a problem in SC-Math6}
\label{fig:example_math_2}
\end{figure*}

\end{document}